\documentclass[journal]{IEEEtran}
\usepackage{leftidx}
\usepackage[pdftex]{graphicx}
\usepackage[cmex10]{amsmath}
\usepackage{epstopdf}
\usepackage{color}
\usepackage{multirow}
\usepackage{booktabs}
\usepackage{subfigure}
\usepackage{url}
\usepackage{balance}
\usepackage{hyperref}
\usepackage{adjustbox,lipsum}

\begin{document}
\title{Segmentation of retinal cysts from Optical Coherence Tomography volumes via selective enhancement}

\author{Karthik Gopinath$^*$,
        Jayanthi Sivaswamy,~\IEEEmembership{Member,~IEEE}
\thanks{*K.Gopinath is with the Centre for Visual Information Technology, International Institute of Information Technology Hyderabad, Hyderabad 500032,India (e-mail: karthik.g@research.iiit.ac.in).}
\thanks{J. Sivaswamy is with the Centre for Visual Information Technology, International Institute of Information Technology Hyderabad, Hyderabad 500032, India (e-mail: jsivaswamy@iiit.ac.in).}
\thanks{}}

\maketitle

\begin{abstract}
Automated and accurate segmentation of cystoid structures in Optical Coherence Tomography (OCT) is of interest in the early detection of retinal diseases. It is however a challenging task. We propose a novel method for localizing cysts in 3D OCT volumes. The proposed work is biologically inspired and based on selective enhancement of the cysts, by inducing motion to a given OCT slice. A Convolutional Neural Network (CNN) is designed to learn a mapping function that combines the result of multiple such motions to produce a probability map for cyst locations in a given slice. The final segmentation of cysts is obtained via simple clustering of the detected cyst locations. The proposed method is evaluated on two public datasets and one private dataset. The public datasets include the one released for the OPTIMA Cyst segmentation challenge (OCSC) in MICCAI 2015 and the DME dataset. After training on the OCSC train set, the method achieves a mean Dice Coefficient (DC) of 0.71 on the OCSC test set. The robustness of the algorithm was examined by cross validation on the DME and AEI (private) datasets and a mean DC values obtained were 0.69 and 0.79, respectively. Overall, the proposed system outperforms all benchmarks. These results underscore the strengths of the proposed method in handling variations in both data acquisition protocols and scanners. 

\end{abstract}


\section{Introduction}

\IEEEPARstart{O}{ptical} Coherence Tomography (OCT) is an imaging modality useful for assessing morphological changes in the sub-retinal layers. Low image quality and high acquisition time hindered the use of OCT in the early days. The newer Spectral-Domain OCT (SD-OCT) overcomes these problems and aids the assessment of small pathological changes. SD-OCT scans can be obtained from multiple vendors at varying resolutions and scanning patterns. Inter scanner variability is reflected in terms of intensity variation and SNR.
Automated detection however has a few challenges: a)Inter-scanner and intra-scanner variability in images (voxel intensities), b) noisy images, c) ambiguous demarcation between sub-retinal layers and finally d) variable shape, size, appearance and arbitrary locations of cysts in the images. Early work on SD-OCT image analysis focused on the intra-retinal tissue layer segmentation \cite{cabrera2005automated} \cite{garvin2008intraretinal} \cite{chiu2010automatic} whereas  attention to abnormality (such as cyst) detection is very recent, with cyst segmentation being posed as a challenge problem in MICCAI 2015.
Cystoid Macular Edema (CME) is a condition in which fluid filled regions (cysts) are formed in the macula leading to swelling \cite{rotsos2008cystoid} \cite{scholl2010pathophysiology}. It is a common retinal disorder that co-occurs in a variety of conditions. If CME is left untreated for more than 6 to 9 months it can lead to chronic macular changes with permanent impairment of central vision. Therefore, the automated detection of CME from SD-OCT images is of interest. 
Automated detection however has a few challenges: a)Inter-scanner and intra-scanner variability in images (voxel intensities), b) noisy images, c) ambiguous demarcation between sub-retinal layers and finally d) variable shape, size, appearance and arbitrary locations of cysts in the images. Early work on SD-OCT image analysis focused on the intra-retinal tissue layer segmentation \cite{cabrera2005automated} \cite{garvin2008intraretinal} \cite{chiu2010automatic} whereas  attention to abnormality (such as cyst) detection is very recent, with cyst segmentation being posed as a challenge problem in MICCAI 2015. 

An early approach proposed to detect cysts was semi-automatic, based on GVF-snake to delineate the intra-retinal and sub-retinal fluid regions in 2D OCT B-scans \cite{fernandez2005delineating}. Later methods were fully automatic and based on 2D analysis. A denoising step was typically followed by a variety of approaches including thresholding and boundary tracing \cite{wilkins2012automated} and texture-based classification of regions of interest derived with a watershed algorithm \cite{gonzalez2013automatic}. 3D methods have been reported more recently. A graph-search/graph-cut based approach that simultaneously segments upper retinal surface, lower retinal surface and fluid-filled regions in \cite{chen2012three} reports good segmentation only for large cysts. Since smaller fluid regions are also of interest, voxel classification based approaches have been attempted in \cite{xu2015stratified} \cite{bogunovic2015geodesic}\cite{chiu2015kernel}. Among the recent approaches are layer-dependent stratified sampling strategy to address the class-imbalance (far fewer cyst vs healthy tissue voxels) \cite{xu2015stratified}, geodesic graph cut method \cite{bogunovic2015geodesic} and a kernel regression-based classification \cite{chiu2015kernel}. Some of these can segment both the layers and fluid-filled abnormalities while others focus only on the abnormalities. 

The MICCAI 2015 OPTIMA cyst segmentation challenge was attempted by participants with a variety of approaches. These range from machine learning with handcrafted features such as intensity and layer thickness, CNN-based voxel labeling \cite{venhuizen2016fully} and graph-based method with layer-dependent information \cite{oguz2016optimal}. The latter method has also been evaluated on the public access DME dataset \cite{dme_datasite}. On the same dataset, a fully convolutional deep learning architecture \cite{roy2017relaynet} is proposed for segmentation of retinal layers and cysts.
The preceding methods are vendor independent. Vendor-specific methods include 3D curvelet transform based dictionary learning \cite{esm} and a marker controlled topographical watershed algorithm \cite{girish2016automated}. Our simple approach \cite{gopinath2016domain} to extract candidates with centre-surround filters and classify with an ensemble random forest failed to perform well on the test set. The proceedings of the challenge can be found at Optima website \cite{optima_url}. In this paper, we propose an entirely different (from existing methods, including ours) approach based on biologically inspired motion patterns and deep learning.

To summarise the current scenario, graph based segmentation such as \cite{chen2012three} is inadequate to segment smaller fluid filled regions which is addressed by existing voxel classification based methods which rely on accurate layer segmentation to derive features. However, segmentation of layers in the presence of abnormalities, is still an open problem. The CNN based method in \cite{venhuizen2016fully} employs three separate networks at different scales to handle the size variability of cysts, which is a major drawback. The 3-network solution increases the time for training and testing and is also memory consuming. The parametric approaches such as \cite{esm} and \cite{girish2016automated} are highly vendor dependent. An alternate approach could be to directly detect the cystoid regions and then segment them. We take such an approach and propose a novel, biologically inspired pipeline for segmentation of cysts. The novel aspects of the proposed method are as follows. 
\begin{itemize}
\item Unique representation for the OCT data, based on motion patterns proposed earlier by our group \cite{deepak2012detection}\cite{deepak2012automatic}.
\item Selective enhancement of the objects of interest (cysts) with a Convolutional Neural Network (CNN).
\item Combining both 2D and 3D information in CNN for detecting the cyst regions.
\item Vendor independent and robust system for detecting and localizing cysts.
\end{itemize}

It is noteworthy that the use of a CNN for enhancing the objects of interest is a major departure from its popular use in literature, namely voxel classification. We show that this novel strategy leads to superior segmentation performance even with a simple, clustering-based post processing step. The next section provides details of the proposed system.

\begin{figure}[htp]
  \centering
  \subfigure[A sample ROI.]{\includegraphics[scale=0.35]{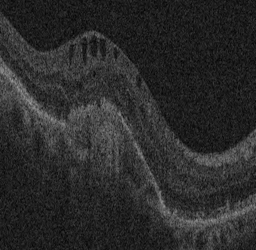}}\quad
  \subfigure[A ROI after denoising]{\includegraphics[scale=0.35]{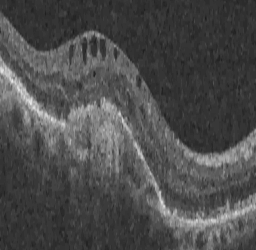}} \quad
  \subfigure[Extracted mask representing layers of interest between ILM and RPE.]{\includegraphics[scale=0.35]{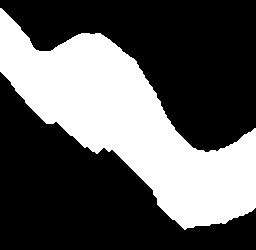}} \quad
\caption{Denoising and extraction of layers of interest on a sample ROI.}
\label{fig:prepro}
\end{figure}

\section{Proposed pipeline}
The proposed pipeline consists of three main stages, namely, preprocessing, detection and clustering. The preprocessing stage is described in sub-section \ref{sec:prep}; the detection of cyst regions via a selective enhancement operation is explained in sub-section \ref{sec:dete} and segmentation via clustering is described in sub-section \ref{sec:segm}. 
\subsection{Preprocessing}
\label{sec:prep}
\subsubsection{Region of interest extraction}
SD-OCT volumes are generally acquired with different operator-defined protocols. Consequently the layer orientation, resolution and number of slices in the acquired volumes vary widely. In order to address this variability, the size of each slice $f(x,y)$ was standardized to a 512$\times$256 pixels by resizing. Since, the region of interest is only 40\% of each slice, for every given volume, a rough region of interest (ROI) was derived as follows: each slice image was projected along columns to obtain a 1D profile and a Gaussian was fit to this profile. The mean value of the Gaussian profile corresponds to a row index $x_0$ in the slice. An ROI of size 250$\times$256, centered at $x_0$ was finally extracted and used for further processing.

\subsubsection{De-noising}
The SD-OCT volumes are degraded by speckle noise. Speckle noise is signal dependent and hence depends on the structure of the tissue in an OCT volume. A total variational denoising method \cite{boyd2004convex} was used to denoise all ROIs. This reduces the texture content and results in a smoother image. A sample ROI extracted from an original slice and the corresponding denoised result are shown in Fig \ref{fig:prepro}(a) and Fig \ref{fig:prepro}(b).

\subsubsection{Extraction of layers of interest}
Cysts are known to be restricted to the layers between Internal Limiting Membrane (ILM) and Retinal Pigment Epithelium (RPE). Various segmentation algorithms have been proposed for segmenting retinal layers. Some of these include pixel intensity variation-based ILM and RPE segmentation \cite{fabritius2009automated}, active contour with a two-step kernel-based optimization scheme \cite{garvin2008intraretinal}, complex diffusion filtering with combined structure tensor replacing thresholding \cite{cabrera2005automated}. In this work, an accurate, graph-based segmentation approach \cite{chiu2010automatic} was used to extract \textit{only} the ILM and RPE layers to form a mask (see Fig. \ref{fig:prepro}(c)). This mask serves as a position prior in the detection stage as explained below.

\begin{figure*}[htp]
  \centering
{\includegraphics[width=\textwidth]{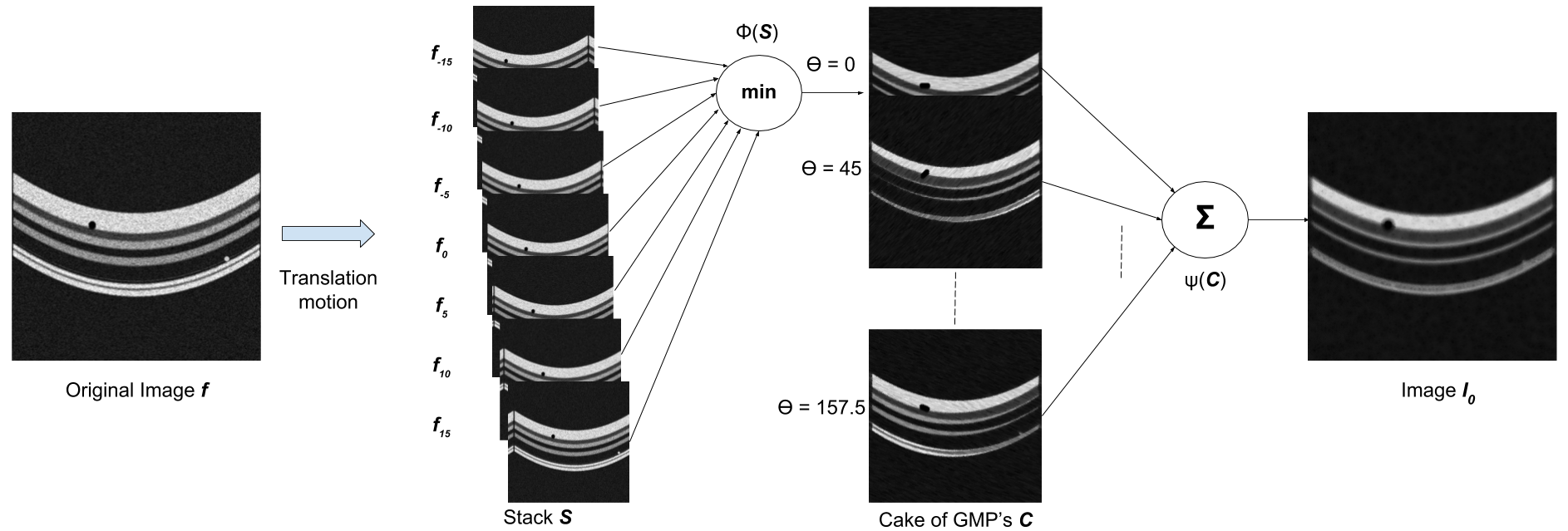}}\quad
  \caption{Understanding generalized motion patterns with a $min$ coalescing function and their sum for a phantom image. }
\label{ph}
\end{figure*}


\subsection{Detection of cyst regions}
Localization of the objects of interest is a vital task in a segmentation problem. We propose to address this task via a selective enhancement process wherein cysts are enhanced without the knowledge of the layer positions. Selective enhancement is done by employing the notion of Generalized Motion Pattern (GMP). The GMP was introduced in our earlier work \cite{deepak2012detection} and has been shown to be effective in diabetic macular edema detection \cite{deepak2012automatic}. In this work, we construct an ensemble of these GMPs and learn a function (using a CNN) to combine them such that only cysts are enhanced. A detailed description is provided next.
\label{sec:dete}
\subsubsection{Generalized motion patterns}

When one captures an image of a racing car, depending on the camera setting, a streak effect or motion blur is observed due to the fast moving object. A related phenomenon is seen when a disc with multi color segments is rotated at an appropriate rate: the colors smear and leave an (visual) impression of a white disc.
GMP is the synthetic (discrete) equivalent of this phenomenon. It is derived by inducing motion (such as translation, rotation) to an image and can be used to represent an image. Let an image be denoted by $f$. Applying motion (transformation T) to $f$ results in a stack $S$ of images 
\begin{equation}
S = \big\{f_j\big\}_{j=-N}^{N}
\end{equation}
where,
\begin{equation}
f_j = f(jT(\delta)); j \in [-N,N]
\end{equation}
where $\delta$ denotes the transformation parameter. In our work, $T$ was chosen to be a translation which is applied in the direction $\theta$, in steps of $\delta$ resulting in an image stack of size $2N+1$. Applying a coalescing function $\phi(.)$ to the stack yields a GMP. 
\begin{equation}
G_{\theta}(f) = \phi(S)= \phi\big\{(f_j)\big\}_{j=-N}^{N}
\end{equation}
The direction of the translation $\theta$ and the choice of coalescing function is dependent on the problem being solved. In the problem at hand, since the object of interest is a cyst which is a dark structure, a minimum ($min$) operation is an appropriate choice for $\phi$. 

In our earlier work, GMPs derived using translation and rotation were shown to be useful in discriminating between \textit{images} of normal and abnormal cases \cite{deepak2012detection}\cite{deepak2012automatic}. The segmentation of abnormalities was not addressed. We propose to solve the cyst segmentation problem by deriving GMPs in multiple directions $\theta$ and combining them such that cysts are selectively enhanced. 

A cake of GMPs ($C$) is constructed by stacking multi-direction GMPs. $C$ is defined as
\begin{equation}
C = \big\{G_{\theta}(f)\big\}^K_{\theta=1} = \bigg\{\phi\big\{(f_j)\big\}_{j=-N}^{N}\bigg\}^K_{\theta=1}
\end{equation}
We propose to map the stack to a single image $I_o$ in which the cysts alone are enhanced. Let $\psi$ denote this mapping. 
\begin{equation}
I_o = \psi(C)
\end{equation}
The above idea is illustrated with a phantom image designed to model an OCT slice in Fig. \ref{ph}. It has a dark blob in the first layer and a bright blob in the last layer representing a cyst and a drusen, respectively. A stack $S$ of translated images obtained from original image is shown in Fig.\ref{ph}. The GMP was derived by applying a $min$ coalescing function on $seven$ translated images $f_j$. A cake of GMP $C$ is obtained by applying translation in eight different $\theta$ directions. It can be observed that the dark blob gets smeared or extended in the layer and the relatively dark layers are wider in the direction $\theta$; whereas the bright blob disappears. The final map $I_o$ is shown in Fig.\ref{ph} with $\psi$ chosen as a simple sum. Here, both morphological changes as well as blur is evident. With a simple $\sum$ for $\psi$, no contextual or shape information is exploited. Hence, there is not much differentiation between a blob and layer in terms of the net effect. This motivates us to \textit{learn} a function that enhances only objects of interest by using relevant information while suppressing other tissues. This type of enhancement is useful when the aim is to localize cysts as it is now possible to interpret the pixel values in the enhanced image as a likelihood (probability) measure for the pixel belonging to a cyst. We propose to use a CNN to learn the function $\psi$ such that cysts are enhanced while other normal anatomical structures are suppressed. 
 


\begin{figure*}
  \centering
    \includegraphics[width=\textwidth]{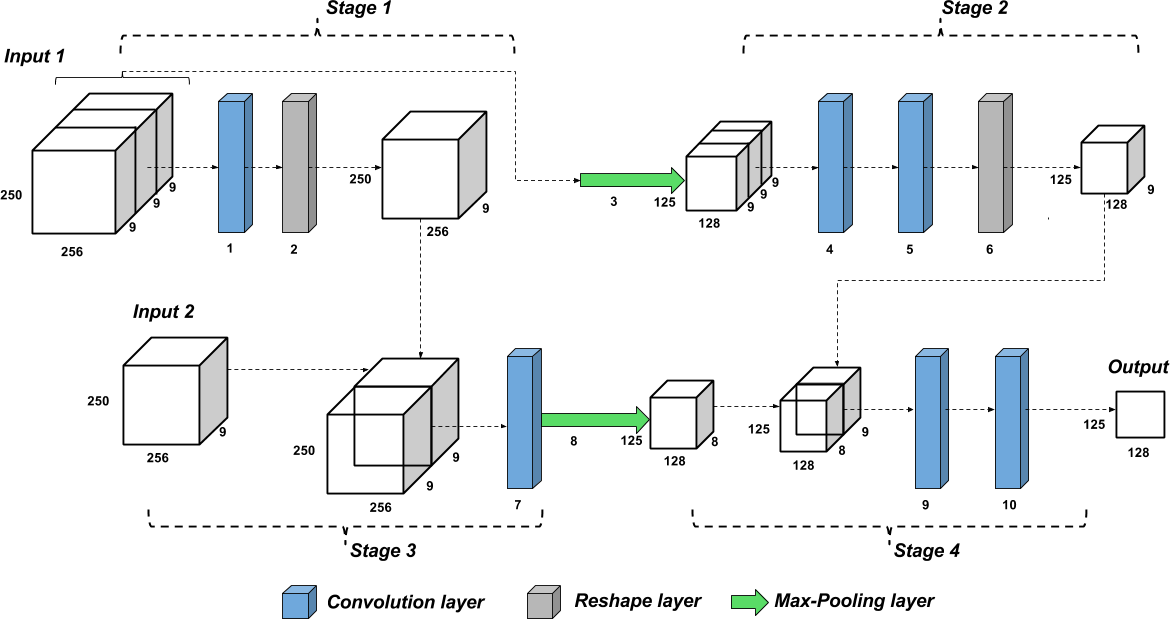}
  \caption{Proposed CNN architecture}
  \label{fig:cnn}
\end{figure*}

\subsubsection{Learning function $\psi$ using CNN}
CNNs are biologically-inspired variants of feed forward multi layer perceptron. Since convolution filter weights are shared across all spatial positions, the free parameters being learned reduces and thus reducing the memory requirements to run the network. A CNN architecture is formed by a stack of layers that transform the input to an output. The three building blocks in our CNN are: (i) convolutional layer with filters that learn specific type of feature in the input at some spatial position  (ii) maxpooling which downsamples the image by retaining the maximum in a local neighborhood (iii) sigmoid activation function that non-linearly maps the intensity values of an image to an interval [0,1].

The task is to detect cysts from a ROI volume of size $\text{X}\times\text{Y}\times\text{Z}$ where $X=250$ and $Y=256$. We utilize the domain knowledge for our work i.e, cysts are 3D structures and thus use consecutive slices as 3D information. The position prior for the cysts is given by the mask extracted from the ROI. In order to detect cysts from the $Z^{th}$ slice, 2 types of inputs (to the CNN) are derived from them. Input1 provides 3D information and hence is a concatenation of the  cakes $C$ (of 250 $\times$ 256 $\times$ K; $K=8$) for the $Z^{th}$ slice and its 2 neighbors, i.e., $(Z-1)^{th}$ and $(Z+1)^{th}$ slices along with their respective position prior. Thus the size of Input1 is 250 $\times$ 256 $\times$ 3 $\times$ 8+1) to CNN. Input2 provides just 2D information for the $Z^{th}$ slice and hence is composed of its Cake $C$ along with its position prior. The size of Input2 is hence 250 $\times$ 256 $\times$ 8+1. All images were normalized to have zero mean value and unit variance. A detailed description of the architecture is given in Table \ref{desc_acri}. Sigmoid activation function is applied on the learned feature map to obtain a probability map.

\begin{table*}[t]
\centering
\caption{Description of the CNN architecture}
\label{desc_acri}
\begin{tabular}{||c|c|c|c|c|c|c|c|c||}
\hline
Input             & Stage       & \# channels & Filter size & \# filters & Pooling size & \begin{tabular}[c]{@{}c@{}}Spatial\\ input size\end{tabular} & \begin{tabular}[c]{@{}c@{}}Spatial\\ output size\end{tabular} & Output          \\ \hline \hline
Input 1           & Convolution & 27          & 1x1x3       & 9          & -           & 250x256x3x9                                                  & 250x256x1x9                                                   & Output 1        \\ 
Output 1          & Reshape     & -           & -           & -          & -           & 250x256x1x9                                                  & 250x256x9                                                     & Output 2        \\ 
Input 1           & Max-pooling & 27          & -           & -          & 2x2x1       & 250x256x3x9                                                  & 125x128x3x9                                                   & Output 3        \\ 
Output 3          & Convolution & 27          & 25x25x3     & 9          & -           & 125x128x3x9                                                  & 125x128x3x9                                                   & Output 4        \\ 
Output 4          & Convolution & 9           & 1x1x3       & 9          & -           & 125x128x3x9                                                  & 125x128x1x9                                                   & Output 5        \\ 
Output 5          & Reshape     & -           & -           & -          & -           & 125x128x1x9                                                  & 125x128x9                                                     & Output 6        \\ 
Input 2;Output 2  & Convolution & 18          & 10x10       & 8          & -           & 250x256x18                                                   & 250x256x8                                                     & Output 7        \\ 
Output 7          & Max-pooling & 27          & -           & -          & 2x2         & 250x256x8                                                    & 125x128x8                                                     & Output 8        \\ 
Output 8;Output 6 & Convolution & 17          & 25x25       & 8          & -           & 125x128x17                                                   & 125x128x8                                                     & Output 9        \\ 
Output 9          & Convolution & 8           & 1x1         & 1          & -           & 125x128x8                                                    & 125x128                                                       & Probability map \\ \hline \hline
\end{tabular}
\end{table*}

The CNN architecture shown in \ref{fig:cnn} is designed to utilize neighborhood information. Stage 1 in the CNN architecture collapses the 3D information into a 2D map without any contextual information. In stage 2 at a lower scale, a large field of view is assessed to combine 3D information to a 2D map. The output of stage 1 is stacked with the second input(Input2) for Stage 3 which enhances the boundary information at the original scale. For the final stage, output from stage2 and stage 3 are concatenated as to form the input. A large neighborhood in 2D is considered for this stage to strengthen the object of interest and weaken the background information. The size of the receptive field is chosen such that the largest cyst in the dataset is enclosed in the field.

\begin{figure*}[htp]

\centering 

  \subfigure[ROI input]{\includegraphics[scale=0.4]{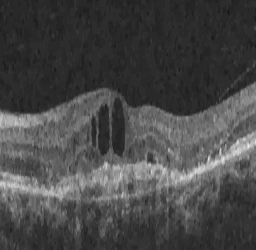}}\quad
  \subfigure[Ground truth: Grader1 $\cap$ Grader2]{\includegraphics[scale=0.4]{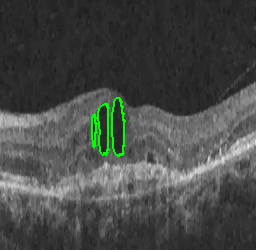}}\quad
  \subfigure[Probability map at $10^{th}$ epoch]{\includegraphics[scale=0.38]{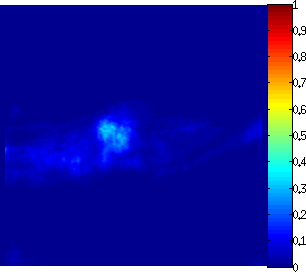}} \quad
  \subfigure[Probability map at $50^{th}$ epoch]{\includegraphics[scale=0.38]{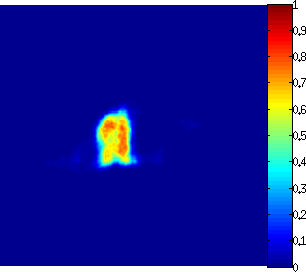}} \quad
  \subfigure[Probability map at $100^{th}$ epoch]{\includegraphics[scale=0.38]{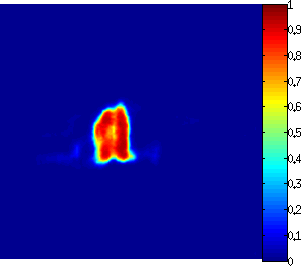}} \quad  
  \subfigure[Probability map at $200^{th}$ epoch]{\includegraphics[scale=0.38]{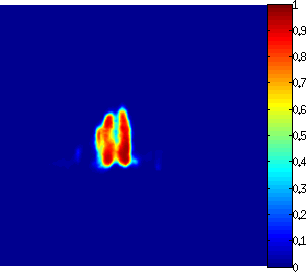}}\quad
  \subfigure[Probability map at $300^{th}$ epoch]{\includegraphics[scale=0.38]{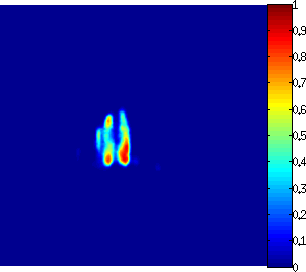}}\quad
  \subfigure[Probability map at $500^{th}$ epoch]{\includegraphics[scale=0.38]{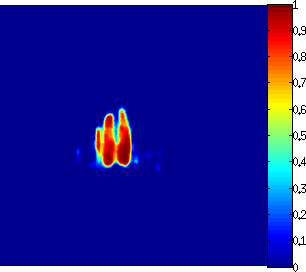}} \quad
  \subfigure[Probability map at $900^{th}$ epoch]{\includegraphics[scale=0.38]{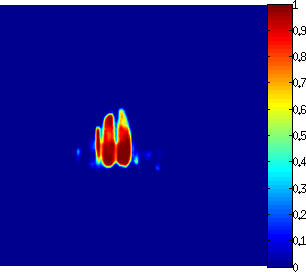}} \quad
  \subfigure[Probability map at $1420^{th}$ epoch]{\includegraphics[scale=0.38]{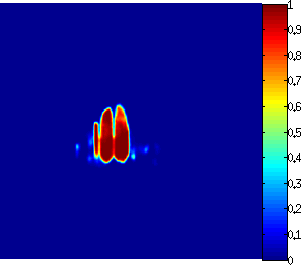}} \quad 
   
\caption{Evolution of the mapping function across epochs. The output of CNN is shown as a probability (heat) map.}
  \label{fig:epc_pmap}
\end{figure*}

A weighted binary cross entropy loss function was chosen to handle the class imbalance in the data, i.e, cyst versus background pixel. This was minimized using gradient descent. The evolution of the probability maps learned (depicted as heat maps), as epochs increase is shown in Fig. \ref{fig:epc_pmap}. It can be seen that cysts gradually take shape in the map.
%

\subsection{Segmentation}
\label{sec:segm}
The trained CNN model learns a function $\psi$ that combines the cake of GMP $C$ to produce a probability map with cyst pixels having higher probability score than background pixels. Due to the smearing effect of GMP, a rough region around the objects of interests (cysts), rather than accurate cyst regions, are detected. Thus, segmentation is required to obtain a precise cyst boundary. The probability map is thresholded to obtain a binary map representing detected cyst regions. This map is multiplied with the ROI image to extract the detected regions in the intensity space. $K$-means clustering is applied on this product image. Detected regions are clustered into cysts, false positives region and background in the intensity space. Since cysts are relatively darker regions compared to the false positive regions, the clusters with lower mean intensity are retained as desired cyst segments.

\section{Experiments and Results}
\begin{table}
\centering
\caption{OCSC dataset description}
\label{data}
\begin{tabular}{||c|c|c|c|c||}
\hline
Scanner  & Cirrus & Nidek & Spectralis & Topcon \\ \hline \hline
Training & 4      & 3     & 4          & 4      \\ 
Testing  & 4      & 3     & 4          & 4      \\ \hline \hline
\end{tabular}
\end{table}

The proposed pipeline was implemented on a NVIDIA GTX 960 GPU, with 4GB of GPU RAM on a core i3 processor. In order to assess the performance of our method, the experiments were carried out on 3 datasets: the MICCAI 2015's OPTIMA Cyst Segmentation Challenge (OCSC) dataset \cite{wu2016multivendor}, the publicly available DME dataset \cite{dme_datasite} and a locally sourced (from Anand Eye Institute, Hyderabad) dataset referred to as AEI. 

The OCSC dataset has 15 training and 15 testing volumes acquired from 4 different scanners; included are manual markings of cysts from two graders. An overview of the dataset is shown in Table \ref{data}. The resolution and density of the volumes vary from $496 \times 512$ to $1024 \times 512$ and 5 to 200 B-scans, respectively. The DME dataset has 10 Spectralis scans with 61 B-scans and $496 \time 768$ resolution each. The manual tracings (from two experts) for these data are provided sparsely (11 B-scans per volume) and not for continuous slices. The AEI dataset has 10 volumes from Optovue scanner with varying number of (30-51) B-scans and fixed resolution of $640 \times 304$. Cyst boundaries for every fifth B-scan of each volume was collected from an expert.

The proposed system was trained on the OCSC training dataset by taking the intersection of the markings of grader 1(G1) and grader 2(G2) as ground truth (GT). Only slices \textit{with} cysts were used during training. A separate validation set was formed by randomly selecting a cyst containing slice from each volume in the OCSC train dataset. The CNN was implemented in Theano using the Keras library. Training for 1500 epochs took about a day and a half. A stochastic gradient descent optimizer as used to minimize the weighted binary cross entropy loss. The training parameters were: learning rate of 0.001; Nesterov momentum was set to 0.75 and batch size was chosen as 8.



\begin{figure*}[htp]
\begin{adjustbox}{minipage=\linewidth,width=\linewidth,height=\hsize/3,scale=0.5}
  \centering
{\includegraphics[width=30mm,height=30mm]{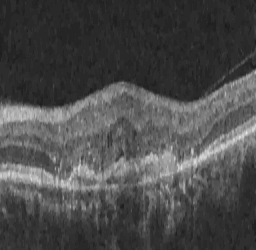}}\quad
{\includegraphics[width=30mm,height=30mm]{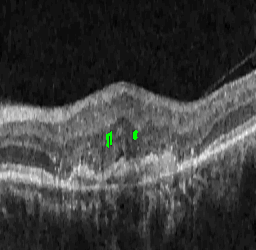}}\quad
{\includegraphics[width=30mm,height=30mm]{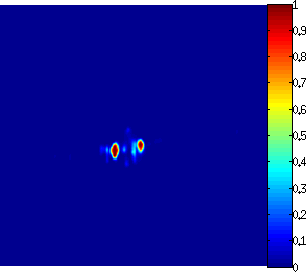}}\quad
{\includegraphics[width=30mm,height=30mm]{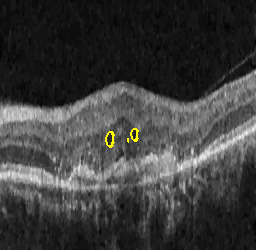}}\quad
{\includegraphics[width=30mm,height=30mm]{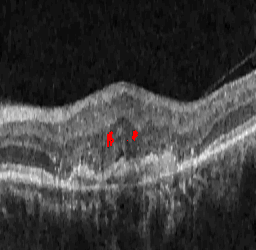}}\vspace{3mm}\quad

{\includegraphics[width=30mm,height=30mm]{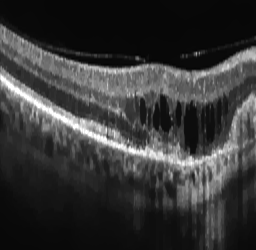}}\quad
{\includegraphics[width=30mm,height=30mm]{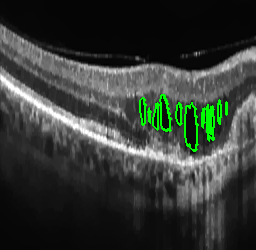}}\quad
{\includegraphics[width=30mm,height=30mm]{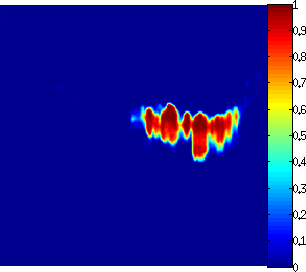}}\quad
{\includegraphics[width=30mm,height=30mm]{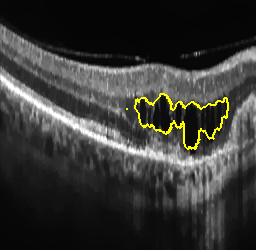}}\quad
{\includegraphics[width=30mm,height=30mm]{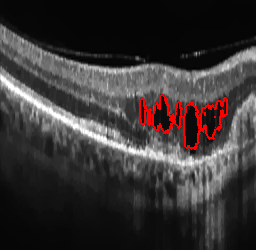}}\vspace{3mm}\quad

{\includegraphics[width=30mm,height=30mm]{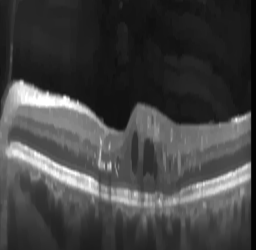}}\quad
{\includegraphics[width=30mm,height=30mm]{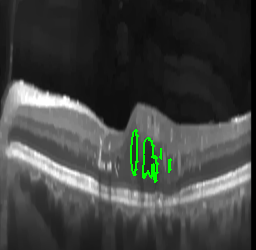}}\quad
{\includegraphics[width=30mm,height=30mm]{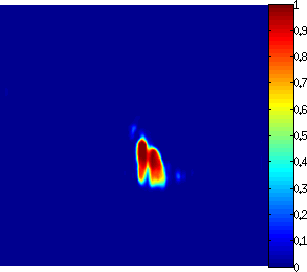}}\quad
{\includegraphics[width=30mm,height=30mm]{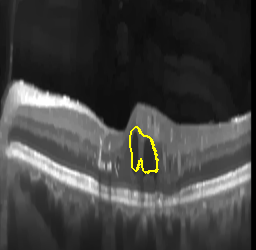}}\quad
{\includegraphics[width=30mm,height=30mm]{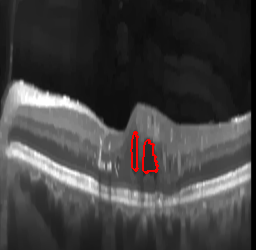}}\vspace{3mm}\quad

\subfigure[ROI of a slice]{\includegraphics[width=30mm,height=30mm]{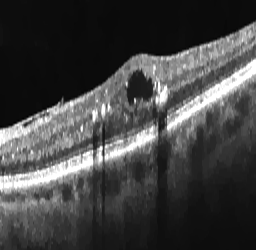}}\quad
\subfigure[Ground truth Grader1 $\cap$ Grader2]{\includegraphics[width=30mm,height=30mm]{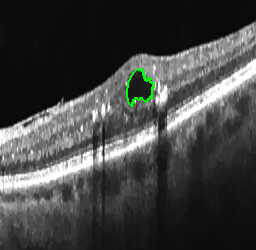}}\quad
\subfigure[Probability map]{\includegraphics[width=30mm,height=30mm]{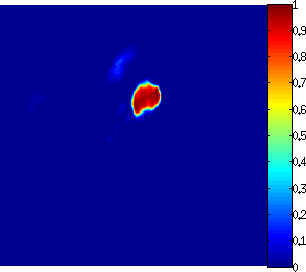}}\quad
\subfigure[Segmentation results by thresholding probability map]{\includegraphics[width=30mm,height=30mm]{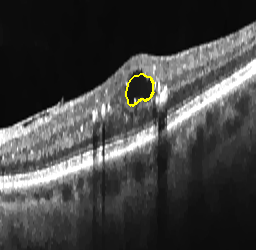}}\quad
\subfigure[Segmentation results using K-means clustering]{\includegraphics[width=30mm,height=30mm]{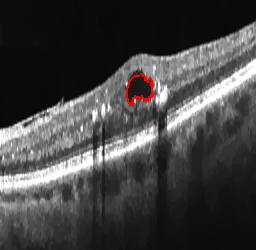}}\quad


\caption{Qualitative results for a slice with cysts. Top to bottom rows: Results for a slice from Cirrus, Nidek, Spectralis, and Topcon scanners.}
\label{withcyst}
 \end{adjustbox}
\end{figure*}

\begin{figure*}[htp]

\begin{adjustbox}{minipage=\linewidth,width=\linewidth,height=\hsize/3,scale=0.5}
\centering 
{\includegraphics[width=30mm,height=30mm]{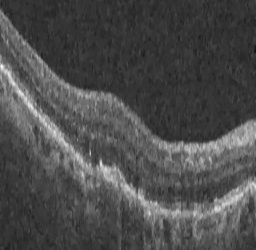}}\quad
{\includegraphics[width=30mm,height=30mm]{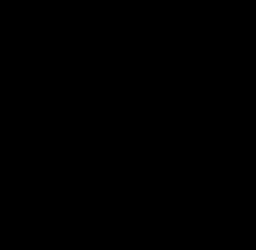}}\quad
{\includegraphics[width=30mm,height=30mm]{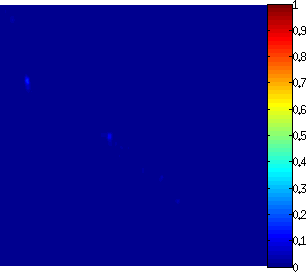}}\quad
{\includegraphics[width=30mm,height=30mm]{qua_res/i5.png}}\quad
{\includegraphics[width=30mm,height=30mm]{qua_res/i5.png}}\vspace{3mm}\quad

{\includegraphics[width=30mm,height=30mm]{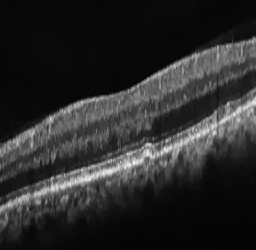}}\quad
{\includegraphics[width=30mm,height=30mm]{qua_res/g8.png}}\quad
{\includegraphics[width=30mm,height=30mm]{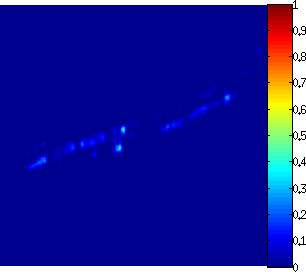}}\quad
{\includegraphics[width=30mm,height=30mm]{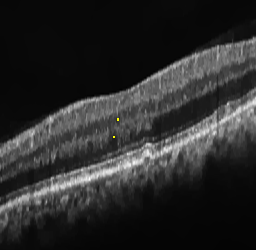}}\quad
{\includegraphics[width=30mm,height=30mm]{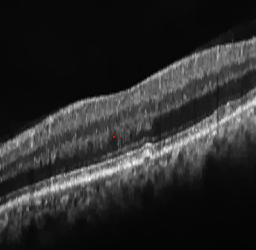}}\vspace{3mm}\quad

{\includegraphics[width=30mm,height=30mm]{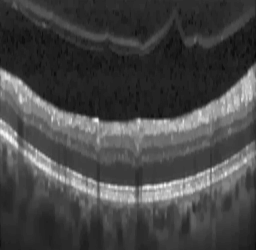}}\quad
{\includegraphics[width=30mm,height=30mm]{qua_res/g8.png}}\quad
{\includegraphics[width=30mm,height=30mm]{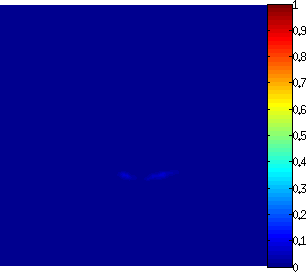}}\quad
{\includegraphics[width=30mm,height=30mm]{qua_res/i7.png}}\quad
{\includegraphics[width=30mm,height=30mm]{qua_res/i7.png}}\vspace{3mm}\quad

\subfigure[ROI of a slice]{\includegraphics[width=30mm,height=30mm]{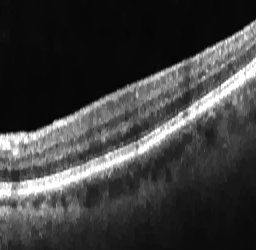}}\quad
\subfigure[Ground truth Grader1 $\cap$ Grader2]{\includegraphics[width=30mm,height=30mm]{qua_res/g8.png}}\quad
\subfigure[Probability map]{\includegraphics[width=30mm,height=30mm]{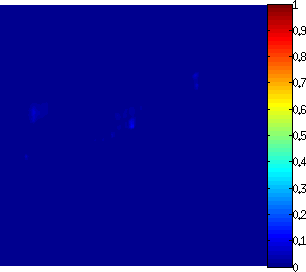}}\quad
\subfigure[Segmentation results by thresholding probability map]{\includegraphics[width=30mm,height=30mm]{qua_res/i8.png}}\quad
\subfigure[Segmentation results using K-means clustering]{\includegraphics[width=30mm,height=30mm]{qua_res/i8.png}}\quad

\caption{Qualitative results for a slice without cyst. Top to bottom rows: Results from a slice from Cirrus, Nidek, Spectralis, and Topcon scanners.}
\label{nop}
 \end{adjustbox}
\end{figure*}
\subsection{Experiments on system design}
Next, we describe a set of experiments that were done to study the impact on system performance. Specifically, these were  to study the effect of i) varying the GMP parameters, ii) variants of the proposed system  and iii) the choice of threshold value which is used to binarise $I_o$. Since OCSC is the largest dataset, all experiments were done by training on the OCSC training set and testing on the OCSC testing set. System performance was assessed by computing the Dice Coefficient (DC) defined below. 
\begin{equation}
Dice \  Coefficient=2{\frac{|Detected \cap  GT| }{|Detected| + |GT|}}
\end{equation}
where $|.|$ denotes the set size. DC$\in$[0,1] can be viewed as a similarity measure over the two sets. DC=1 is the ideal value as it indicates a perfect match between computed result and GT. 

\subsubsection{Effect of different GMP parameters}
\begin{figure}
  \centering
    \includegraphics[width=0.4\textwidth]{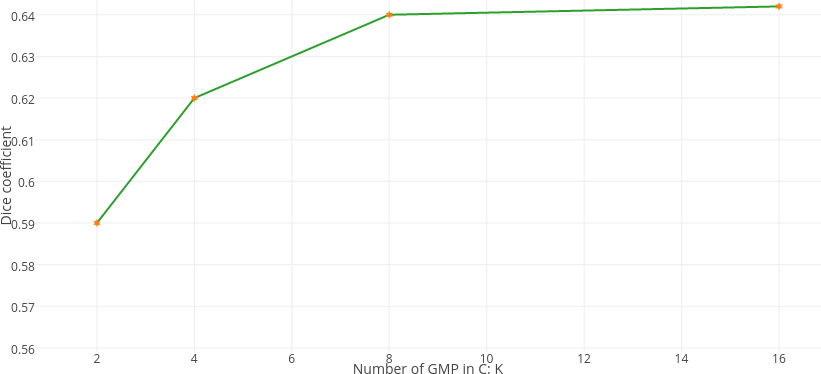}
  \caption{Segmentation performance as a function of the number ($K$) of GMPs.}
  \label{fig:the}
\end{figure}

GMP computation requires choosing a few parameters such as the step size $\delta$, range or extent of translation $N$ and the number of translation directions or number of GMPs $K$. The last two directly affect the information content and size of cake $C$ which is mapped by the CNN (by learning a function). $K$ and $N$ were varied while $\delta$ was held constant ($=1$). The output of the CNN ($I_o$) was thresholded to obtain the detected cyst regions. Since these are highly localized, they were treated as roughly segmented cysts and evaluated using DC. Fig. \ref{fig:the} shows the computed DC as a function of $K$ with $N=5$. 
  
\begin{figure}
  \centering
    \includegraphics[width=0.4\textwidth]{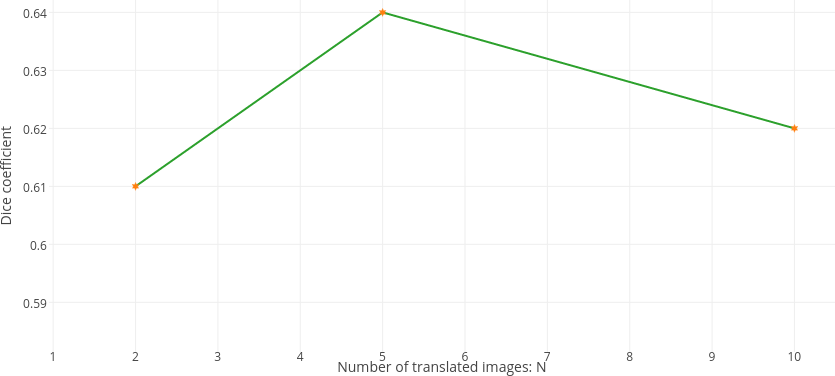}
  \caption{Segmentation performance as a function of the extent ($N$) of translation in each GMP.}
  \label{fig:smear}
\end{figure}

The DC is seen to increase initially and saturate after $K = 8$. For $K < 8$ the cake is smaller which implies has less information to CNN and hence inadequate learning. Since the training time increases with the cake size, $K=8$ was chosen for all our further experiments.

Next, the extent of translation $N$ was varied. Specifically, $N=2,5,10$ were considered. Small (large) value for $N$ implies a small (large) range of induced motion. Since motion serves to spatially extend the cyst (along $\theta$), a large $N$ is potentially beneficial for detection of small cysts. However, this can lead to merging of cysts with dark background or give rise to false positives, given the fairly dense retinal layer structure. A small $N$ helps overcome these problems, however at the risk of failure to detect small cysts. Thus the DC versus $N$ plot can be expected to have a mode. This is confirmed in Fig. \ref{fig:smear}. \\
In experiments described henceforth, each of the GMP was created by applying translation motion on a ROI slice with $N=5$, $\delta =1$ and $\theta$ = $0  ^{\circ}$ to $180^{\circ}$ (in steps of $22.5^{\circ}$); $C$ was derived by using a $min$ coalescing function over a stack of $K=8$ GMPs.

\subsubsection{Effect of variants of the system}
The proposed CNN architecture combines contextual information at 2D as well as 3D levels. To assess their relative effectiveness, a simple CNN architecture which relies only on 2D local information was considered by plugging out stages 1 and 2 in the proposed architecture. A cyst is a 3D structure and hence the 2D CNN model can be expected to perform poorly compared to 2D+3D CNN model (proposed). The DC listed in Table \ref{difcnn_2d3d}, is seen to improve with the addition of 3D information by 14\%. 

\subsubsection{Effect on threshold at the CNN output}
\begin{figure}
  \centering
    \includegraphics[width=0.4\textwidth]{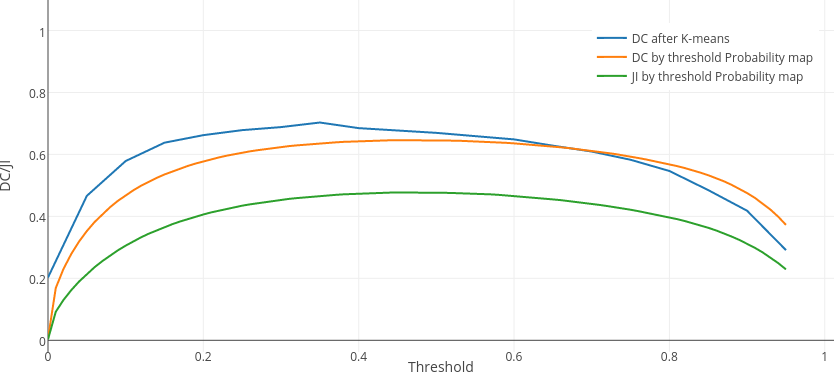}
  \caption{DC and JI for various thresholds at different stages}
  \label{dcji}
\end{figure}

The thresholded result of the probability map (CNN output) represents a rough segmentation. The impact of this threshold on rough and fine segmentation was also assessed. The plot in figure Fig. \ref{dcji} shows the DC and Jaccard index (JI) as a function of threshold level. Both increase initially as the rough segments continue to improve. DC and JI fall with threshold values above $0.5$ since more noisy pixels are passed on. When the threshold exceeds $0.7$ the DC obtained after clustering drops below the DC obtained prior to clustering. This is because, thresholding at a high probability score leads to fewer false positive (FP) regions and under segmentation, which in turn affects clustering. However, at a threshold of $0.35$, clustering is seen to improve the performance by $7.8\% $. This was chosen as the optimal value for system evaluation.
 
\begin{table}[tbp]
\centering
\caption{Mean DC values in comparison with U-Net with data augmentation}
\label{difcnn_2d3d}
\begin{tabular}{||c|c|c|c||}
\hline
Input     & 2D U-Net & Proposed model  & \begin{tabular}[c]{@{}c@{}} Proposed model + K-means clustering \end{tabular} \\ 
\hline \hline
 & 0.56     & 0.64 & \textbf{0.69} \\ \hline \hline
\end{tabular}
\end{table}

\begin{table}[]
\centering
\caption{Mean DC values obtained with CNN model variants}
\label{difcnn}
\begin{tabular}{||c|c|c|c||}
\hline
\multicolumn{1}{||c|}{Method} & \multicolumn{1}{|c|}{\begin{tabular}[c]{@{}c@{}}OCSC \\ Test set\end{tabular}} & \multicolumn{1}{|c|}{\begin{tabular}[c]{@{}c@{}}DME\\ Dataset\end{tabular}} & \multicolumn{1}{|c||}{\begin{tabular}[c]{@{}c@{}}AEI\\ Dataset\end{tabular}} \\ \hline \hline 
\multicolumn{1}{||c|}{U-net with data augmentation} & \multicolumn{1}{|c|}{0.65} & \multicolumn{1}{|c|}{0.58} & \multicolumn{1}{|c||}{0.69} \\ 
Proposed system + K-means & 0.69 & 0.67 & 0.79 \\ \hline \hline
\end{tabular}
\end{table}


\subsection{Evaluation of the system performance}
Qualitative results of the proposed method are shown in Fig. \ref{withcyst} and Fig. \ref{nop}. The probability map computed by the CNN, thresholded result overlaid on the original image and the final segmentation results are shown for a sample slice from 4 different scanners. It can be observed that the method is able to produce probability maps which highlight the cysts when present and nearly flat maps when cysts are absent. The thresholded outputs (in column d of Fig. \ref{withcyst}) and final segments (in column e of Fig. \ref{withcyst}) indicate that segmentation is consistently good for large and small cysts and the false detections are also minimal. The results also appear to be robust to inter scanner variation in contrast/intensity. 


Next, we present the segmentation performance of the proposed system. Metrics such as DC, JI, Positive Predictive Value (PPV) and Sensitivity were used by OCSC organizers for evaluating the different segmentation methods. Since DC and JI convey very similar message, we choose to evaluate our work using DC alone.
In order to assess the strengths of the proposed solution, we evaluate the work quantitatively against the U-Net \cite{ronneberger2015u} a widely used deep network for biomedical image segmentation tasks. It was trained end to end on the OCSC training set with standard augmentation. The obtained mean DC for the OCSC test set and cross validation results on DME and AEI dataset are listed in Table \ref{difcnn}. 
The 6.1\% improvement seen with GMP as input relative to ROI as input, substantiates the importance of using appropriate representation for the input data. U-net has pooling and upsampling stages in the architecture which are memory intensive. The training for U-net had to be done on a GPU (NVIDIA TITAN X) with higher memory capacity.

Finally, we benchmark the proposed system against i) the participants of the challenge on OCSC test set and ii) methods reported on DME dataset. The mean DC values are presented in Table \ref{dice_dataset}. The results on the OCSC test set for the participants of the challenge are taken from the OCSC website \cite{optima_res}. The overall DC values obtained for different datasets are listed (in rank order) separately in Table \ref{dice_dataset}. Here, G1 and G2 denote the GT from 2 graders; Unmasked (U) denotes the entire volume while Masked (M) denotes a volume with masked central 3 mm circular region centered at macula as described in \cite{wu2016multivendor}; (S) in the row of Esmaeili et al. denotes the results are for only Spectralis scanner. It should be noted that results reported in \cite{esm} are for the OCSC training set whereas the DC values listed in the table are for the OCSC test set. The DC values should be nearly equal for M and U volumes if detection is not sensitive to location and this holds only for the proposed method which outperforms all other methods on the OCSC test dataset. 

In the DME dataset, the uncertainty of fluid boundaries is reflected by the difference in DC values (0.65/0.9) for the two manual raters. Both \cite{chiu2015kernel} \cite{oguz2016optimal} have reported results on this dataset though the latter report only the volume similarity error metric and not DC. We consider DC as the standard metric for evaluating the performance of a segmentation task and hence benchmark with \cite{chiu2015kernel}. The DC values against the two experts as well as their union on DME dataset are given in Table \ref{dice_dataset}. On the AEI dataset, the obtained DC value was \textbf{0.79/0.18}. The superior performance of the proposed system on the DME dataset and AEI dataset is noteworthy given that our system was \textit{not} separately trained on these datasets.

\begin{table*}[thbp]
\centering
\caption{Dice coefficient for the OCSC dataset, DME dataset and AEI dataset}
\label{dice_dataset}
\resizebox{\linewidth}{!}{\begin{tabular}{||c||ccc|ccc||ccc||c|}
\hline
\multirow{3}{*}{} & \multicolumn{6}{c||}{OCSC Test dataset} & \multicolumn{3}{c||}{DME dataset} \\  \cline{2-10} 
 & \multicolumn{3}{c|}{U} & \multicolumn{3}{c||}{M} & \multirow{2}{*}{\begin{tabular}[c|]{@{}c@{}}Expert 1 \\ (MJA)\end{tabular}} & \multirow{2}{*}{\begin{tabular}[c]{@{}c@{}}Expert 2\\  (PSM)\end{tabular}} & \multirow{2}{*}{\begin{tabular}[c]{@{}c@{}}Union \\ (MJA , PSM)\end{tabular}} \\ \cline{2-7}
 & G1 & G2 & G1$\cap$G2 & G1 & G2 & G1$\cap$G2 &  &  &  \\  \hline
Proposed Work & \textbf{0.67/0.17} & \textbf{0.68/0.17} & \textbf{0.69/0.18} & \textbf{0.70/0.17} & \textbf{0.70/0.15} & \textbf{0.71/0.16} & 0.69/0.17 & 0.67/0.18 & 0.67/0.17 \\
de Sisternes et al. & 0.64/0.14 & 0.63/0.14 & 0.65/0.15 & 0.68/0.15 & 0.67/0.17 & 0.69/0.15 & - & - & - \\ 
Venhuizen et al.~\cite{venhuizen2016fully} & 0.56/0.2 & 0.55/0.22 & 0.54/0.20 & 0.61/0.19 & 0.60/0.19 & 0.59/0.19 & - & - & - \\ 
Oguz et al.~\cite{oguz2016optimal} & 0.48/0.25 & 0.48/0.22 & 0.48/0.22 & 0.60/0.15 & 0.59/0.15 & 0.60/0.14 & - & - & - \\ 
Esmaeili et al.~\cite{esm} & 0.46/0.25 & 0.45/0.24 & 0.45/0.25 & 0.55/0.27 & 0.55/0.27 & 0.55/0.28 & - & - & - \\
Haritz et al.~\cite{gopinath2016domain} & 0.14/0.08 & 0.14/0.08 & 0.14/0.08 & 0.23/0.15 & 0.23/0.15 & 0.23/0.15 & - & - & - \\ 
\cite{chiu2015kernel} & - & - & - & - & - & - & - & - & 0.53/0.34 \\ 
\cite{roy2017relaynet} & - & - & - & - & - & - & - & - & \textbf{0.77/-} \\ \hline \hline
\end{tabular}}
\end{table*}

\begin{figure}[htp]
\centering 


{\includegraphics[width=20mm,height=20mm]{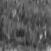}}\quad
{\includegraphics[width=20mm,height=20mm]{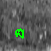}}\quad
{\includegraphics[width=20mm,height=20mm]{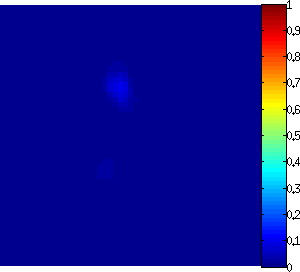}}\vspace{3mm}\quad

{\includegraphics[width=20mm,height=20mm]{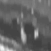}}\quad
{\includegraphics[width=20mm,height=20mm]{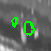}}\quad
{\includegraphics[width=20mm,height=20mm]{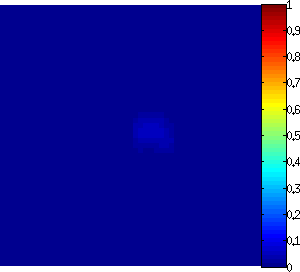}}\vspace{3mm}\quad

\subfigure[A sub image around cyst]{\includegraphics[width=20mm,height=20mm]{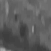}}\quad
\subfigure[Ground truth (Grader1 $\cap$ Grader2)]{\includegraphics[width=20mm,height=20mm]{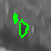}}\quad
\subfigure[Probability map]{\includegraphics[width=20mm,height=20mm]{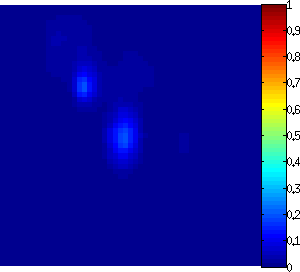}}\quad

\caption{Sample failure (to detect) cases.}
\label{failcnn}
\end{figure}
\section{Discussion and Conclusions}

An automated retinal cyst segmentation algorithm was presented. It is a generic approach which is independent of scanners and does not require layer segmentation which is in contrast to existing approaches. For instance, layer-dependent statistical information is used to define a cost function in the graph based method in \cite{oguz2016optimal} and serves as a feature in machine learning based algorithm in \cite{chiu2015kernel}. 
Our method depends only on extracting an ROI between ILM and RPE for processing. The heart of the proposed method is a novel representation derived by inducing motion. The CNN is used to selectively enhance the object of interest and aid in detection of the object. As only few parameters are to be learnt, CNN training is possible with only 584 sample slices (containing cysts) from the OCSC training dataset. The architecture of the CNN is designed to utilize contextual information in both 2D and 3D and at multiple scales which results in good performance on large cysts and small fluid regions. The effectiveness of the CNN is underscored by the fact that its thresholded output is able to achieve results similar to the other methods proposed during the MICCAI challenge.  A simple clustering on this output improved the DC by nearly $7.8\%$. The cross validation results on the DME dataset and AEI dataset points to the robustness of the proposed method. On explicit training on DME dataset, DC of 0.77 was achieved by \cite{roy2017relaynet}.

The proposed system has some limitations. It fails to detect cyst regions which are very small in a volume especially when they are isolated in depth. These are sometimes treated as noise and rejected. Sample failure cases are shown in Fig. \ref{failcnn}. It also tends to merge small cysts, when they are adjacent to a large one and this in turn affects the DC value. One such case is seen in the second row of Fig. \ref{withcyst}. 

Besides cysts which are fluid filled regions, quantifying fatty deposits such as drusen in the RPE layer, which occur in Age-related macular degeneration, are also of interest in OCT image analysis. The proposed segmentation method can easily be extended to handle drusen by simply replacing the coalescing function with $max$. This we believe is because our strategy for localization is selective enhancement of the abnormality of interest, agnostic to the location of the abnormality. Further improvement in the work is possible by embedding domain-assisted spatial information in the input to the CNN. 

\bibliographystyle{IEEEtran}
\bibliography{ab_short}

\begin{thebibliography}{10}
\providecommand{\url}[1]{#1}
\csname url@samestyle\endcsname
\providecommand{\newblock}{\relax}
\providecommand{\bibinfo}[2]{#2}
\providecommand{\BIBentrySTDinterwordspacing}{\spaceskip=0pt\relax}
\providecommand{\BIBentryALTinterwordstretchfactor}{4}
\providecommand{\BIBentryALTinterwordspacing}{\spaceskip=\fontdimen2\font plus
\BIBentryALTinterwordstretchfactor\fontdimen3\font minus
  \fontdimen4\font\relax}
\providecommand{\BIBforeignlanguage}[2]{{%
\expandafter\ifx\csname l@#1\endcsname\relax
\typeout{** WARNING: IEEEtran.bst: No hyphenation pattern has been}%
\typeout{** loaded for the language `#1'. Using the pattern for}%
\typeout{** the default language instead.}%
\else
\language=\csname l@#1\endcsname
\fi
#2}}
\providecommand{\BIBdecl}{\relax}
\BIBdecl

\bibitem{cabrera2005automated}
D.~Cabrera~Fern{\'a}ndez, H.~M. Salinas, and C.~A. Puliafito, ``Automated
  detection of retinal layer structures on optical coherence tomography
  images,'' \emph{Optics Express}, vol.~13, no.~25, pp. 10\,200--10\,216, 2005.

\bibitem{garvin2008intraretinal}
M.~K. Garvin, M.~D. Abr{\`a}moff, R.~Kardon, S.~R. Russell, X.~Wu, and
  M.~Sonka, ``Intraretinal layer segmentation of macular optical coherence
  tomography images using optimal 3-d graph search,'' \emph{Medical Imaging,
  IEEE Transactions on}, vol.~27, no.~10, pp. 1495--1505, 2008.

\bibitem{chiu2010automatic}
S.~J. Chiu, X.~T. Li, P.~Nicholas, C.~A. Toth, J.~A. Izatt, and S.~Farsiu,
  ``Automatic segmentation of seven retinal layers in sdoct images congruent
  with expert manual segmentation,'' \emph{Optics express}, vol.~18, no.~18,
  pp. 19\,413--19\,428, 2010.

\bibitem{rotsos2008cystoid}
T.~G. Rotsos and M.~M. Moschos, ``Cystoid macular edema,'' \emph{Clinical
  ophthalmology (Auckland, NZ)}, vol.~2, no.~4, p. 919, 2008.

\bibitem{scholl2010pathophysiology}
S.~Scholl, J.~Kirchhof, and A.~J. Augustin, ``Pathophysiology of macular
  edema,'' \emph{Ophthalmologica}, vol. 224, no. Suppl. 1, pp. 8--15, 2010.

\bibitem{fernandez2005delineating}
D.~C. Fernandez, ``Delineating fluid-filled region boundaries in optical
  coherence tomography images of the retina,'' \emph{IEEE transactions on
  medical imaging}, vol.~24, no.~8, pp. 929--945, 2005.

\bibitem{wilkins2012automated}
G.~R. Wilkins, O.~M. Houghton, and A.~L. Oldenburg, ``Automated segmentation of
  intraretinal cystoid fluid in optical coherence tomography,'' \emph{IEEE
  Transactions on Biomedical Engineering}, vol.~59, no.~4, pp. 1109--1114,
  2012.

\bibitem{gonzalez2013automatic}
A.~Gonz{\'a}lez, B.~Remeseiro, M.~Ortega, M.~G. Penedo, and P.~Charl{\'o}n,
  ``Automatic cyst detection in oct retinal images combining region flooding
  and texture analysis,'' in \emph{Proceedings of the 26th IEEE International
  Symposium on Computer-Based Medical Systems}.\hskip 1em plus 0.5em minus
  0.4em\relax IEEE, 2013, pp. 397--400.

\bibitem{chen2012three}
X.~Chen, M.~Niemeijer, L.~Zhang, K.~Lee, M.~D. Abr{\`a}moff, and M.~Sonka,
  ``Three-dimensional segmentation of fluid-associated abnormalities in retinal
  oct: probability constrained graph-search-graph-cut,'' \emph{IEEE
  transactions on medical imaging}, vol.~31, no.~8, pp. 1521--1531, 2012.

\bibitem{xu2015stratified}
X.~Xu, K.~Lee, L.~Zhang, M.~Sonka, and M.~D. Abramoff, ``Stratified sampling
  voxel classification for segmentation of intraretinal and subretinal fluid in
  longitudinal clinical oct data,'' \emph{IEEE transactions on medical
  imaging}, vol.~34, no.~7, pp. 1616--1623, 2015.

\bibitem{bogunovic2015geodesic}
H.~Bogunovi{\'c}, M.~D. Abr{\`a}moff, and M.~Sonka, ``Geodesic graph cut based
  retinal fluid segmentation in optical coherence tomography,'' 2015.

\bibitem{chiu2015kernel}
S.~J. Chiu, M.~J. Allingham, P.~S. Mettu, S.~W. Cousins, J.~A. Izatt, and
  S.~Farsiu, ``Kernel regression based segmentation of optical coherence
  tomography images with diabetic macular edema,'' \emph{Biomedical optics
  express}, vol.~6, no.~4, pp. 1172--1194, 2015.

\bibitem{venhuizen2016fully}
F.~Venhuizen, M.~J. van Grinsven, B.~van Ginneken, C.~C. Hoyng, T.~Theelen, and
  C.~I. Sanchez, ``Fully automated segmentation of intraretinal cysts in 3d
  optical coherence tomography,'' \emph{Investigative Ophthalmology \& Visual
  Science}, vol.~57, no.~12, pp. 5949--5949, 2016.

\bibitem{oguz2016optimal}
I.~Oguz, L.~Zhang, M.~D. Abr{\`a}moff, and M.~Sonka, ``Optimal retinal cyst
  segmentation from oct images,'' in \emph{SPIE Medical Imaging}.\hskip 1em
  plus 0.5em minus 0.4em\relax International Society for Optics and Photonics,
  2016, pp. 97\,841E--97\,841E.

\bibitem{dme_datasite}
\BIBentryALTinterwordspacing
Dataset. Dme dataset from duke univerity. [Online]. Available:
  \url{http://people.duke.edu/~sf59/Chiu_BOE_2014_dataset.htm}
\BIBentrySTDinterwordspacing

\bibitem{roy2017relaynet}
A.~G. Roy, S.~Conjeti, S.~P.~K. Karri, D.~Sheet, A.~Katouzian, C.~Wachinger,
  and N.~Navab, ``Relaynet: Retinal layer and fluid segmentation of macular
  optical coherence tomography using fully convolutional network,'' \emph{arXiv
  preprint arXiv:1704.02161}, 2017.

\bibitem{esm}
R.~H. H.~F. Mahdad Esmaeili~M, Dehnavi~AM, ``Three-dimensional segmentation of
  retinal cysts from spectral-domain optical coherence tomography images by the
  use of three-dimensional curvelet based k‑svd,'' \emph{J Med Sign Sence},
  vol.~6, pp. 166--171, 2016.

\bibitem{girish2016automated}
G.~Girish, A.~R. Kothari, and J.~Rajan, ``Automated segmentation of
  intra-retinal cysts from optical coherence tomography scans using marker
  controlled watershed transform,'' in \emph{Engineering in Medicine and
  Biology Society (EMBC), 2016 IEEE 38th Annual International Conference of
  the}.\hskip 1em plus 0.5em minus 0.4em\relax IEEE, 2016, pp. 1292--1295.

\bibitem{gopinath2016domain}
K.~Gopinath and J.~Sivaswamy, ``Domain knowledge assisted cyst segmentation in
  oct retinal images,'' \emph{arXiv preprint arXiv:1612.02675}, 2016.

\bibitem{optima_url}
\BIBentryALTinterwordspacing
OPTIMA. Proceeding of the miccai optima cyst segmentation challenge. [Online].
  Available: \url{https://optima.meduniwien.ac.at/research/challenges/}
\BIBentrySTDinterwordspacing

\bibitem{deepak2012detection}
K.~S. Deepak, N.~K. Medathati, and J.~Sivaswamy, ``Detection and discrimination
  of disease-related abnormalities based on learning normal cases,''
  \emph{Pattern Recognition}, vol.~45, no.~10, pp. 3707--3716, 2012.

\bibitem{deepak2012automatic}
K.~S. Deepak and J.~Sivaswamy, ``Automatic assessment of macular edema from
  color retinal images,'' \emph{IEEE Transactions on Medical Imaging}, vol.~31,
  no.~3, pp. 766--776, 2012.

\bibitem{boyd2004convex}
S.~Boyd and L.~Vandenberghe, \emph{Convex optimization}.\hskip 1em plus 0.5em
  minus 0.4em\relax Cambridge university press, 2004.

\bibitem{fabritius2009automated}
T.~Fabritius, S.~Makita, M.~Miura, R.~Myllyl{\"a}, and Y.~Yasuno, ``Automated
  segmentation of the macula by optical coherence tomography,'' \emph{Optics
  express}, vol.~17, no.~18, pp. 15\,659--15\,669, 2009.

\bibitem{wu2016multivendor}
J.~Wu, A.-M. Philip, D.~Podkowinski, B.~S. Gerendas, G.~Langs, C.~Simader,
  S.~M. Waldstein, and U.~M. Schmidt-Erfurth, ``Multivendor spectral-domain
  optical coherence tomography dataset, observer annotation performance
  evaluation, and standardized evaluation framework for intraretinal cystoid
  fluid segmentation,'' \emph{Journal of Ophthalmology}, vol. 2016, 2016.

\bibitem{ronneberger2015u}
O.~Ronneberger, P.~Fischer, and T.~Brox, ``U-net: Convolutional networks for
  biomedical image segmentation,'' in \emph{International Conference on Medical
  Image Computing and Computer-Assisted Intervention}.\hskip 1em plus 0.5em
  minus 0.4em\relax Springer, 2015, pp. 234--241.

\bibitem{optima_res}
\BIBentryALTinterwordspacing
OPTIMA. Results of the miccai optima cyst segmentation challenge. [Online].
  Available:
  \url{http://optima.meduniwien.ac.at/challenges/optima-segmentation-challenge-1/results/}
\BIBentrySTDinterwordspacing

\end{thebibliography}

\end{document}